\documentclass{article}

\PassOptionsToPackage{numbers, compress}{natbib}

\usepackage[preprint]{neurips_2022}

\usepackage[utf8]{inputenc} 
\usepackage[T1]{fontenc}    
\usepackage{hyperref}       
\usepackage{url}            
\usepackage{booktabs}       
\usepackage{amsfonts}       
\usepackage{nicefrac}       
\usepackage{microtype}      
\usepackage{xcolor}         

\usepackage{graphicx}
\usepackage{tabularray}
\usepackage{cleveref}
\crefname{section}{\S}{\S}
\crefname{Section}{\S}{\S}
\crefformat{section}{§#2#1#3}
\crefname{table}{Tab.}{Tab.}
\crefname{appendix_table}{Tab.}{Tab.}
\crefname{Table}{Tab.}{Tab.}
\crefname{Figure}{Fig.}{Fig.}
\crefname{figure}{Fig.}{Fig.}
\crefname{appendix}{Appendix}{Appendix}
\crefname{chapter}{Chapter}{Chapter}

\hypersetup{breaklinks=true}

\usepackage{array}
\usepackage{tabularray}
\UseTblrLibrary{booktabs}
\definecolor{red_tab}{rgb}{0.816,0,0}
\definecolor{green_tab}{rgb}{0.227,0.596,0.227}
\SetTblrInner{rowsep=2.5pt}
\usepackage{ragged2e}

\usepackage{amsthm}
\newtheoremstyle{theorem}{4mm}{1mm}{\itshape}{ }{\bfseries}{.}{ }{}
\theoremstyle{theorem}

\newtheorem{theoremIB}{\ignorespaces}

\newtheorem{theoremIR}{\ignorespaces}

\newtheorem{theoremSB}{\ignorespaces}

\newtheorem{theoremSR}{\ignorespaces}

\title{\textit{Personalisation within bounds}: A risk taxonomy and policy framework for the alignment of large language models with personalised feedback}

\author{%
  Hannah Rose Kirk$^{1}$\thanks{Corresponding author: \texttt{hannah.kirk@oii.ox.ac.uk}} \\
  \And
  Bertie Vidgen$^{1}$
    \And
  Paul Röttger$^{1}$
  \And
  Scott A. Hale$^{1}$
  \And
  \\
  ${^1}$ Oxford Internet Institute, University of Oxford
}

\begin{document}

\maketitle

\begin{abstract}
Large language models (LLMs) are used to generate content for an increasingly wide range of tasks, and are set to reach a growing audience in coming years due to integration in product interfaces like ChatGPT or search engines like Bing.
This intensifies the need to ensure that models are aligned with human preferences and do not produce unsafe, inaccurate or toxic outputs. 
While alignment techniques like reinforcement learning with human feedback (RLHF) and red-teaming can mitigate some safety concerns and improve model capabilities, it is unlikely that an aggregate fine-tuning process can adequately represent the full range of users' preferences and values. 
Different people may legitimately disagree on their preferences for language and conversational norms, as well as on values or ideologies which guide their communication. \textit{Personalising} LLMs through micro-level preference learning processes may result in models that are better aligned with each user. However, there are several normative challenges in defining the bounds of a societally-acceptable and safe degree of personalisation. 
In this paper, we ask how, and in what ways, LLMs should be personalised. First, we review literature on current paradigms for aligning LLMs with human feedback, and identify issues including (i) a lack of clarity regarding what alignment means; (ii) a tendency of technology providers to prescribe definitions of inherently subjective preferences and values; and (iii) a ``tyranny of the crowdworker'', exacerbated by a lack of documentation in who we are really aligning to.
Second, we present a taxonomy of benefits and risks associated with personalised LLMs, for individuals and society at large. Finally, we propose a three-tiered policy framework that allows users to experience the benefits of personalised alignment, while restraining unsafe and undesirable LLM-behaviours within (supra-)national and organisational bounds.
\end{abstract}

\section{Introduction}
The capabilities of large language models (LLMs) to complete tasks and follow natural language instructions has substantially improved in recent years \cite{ouyangTraining2022}. LLMs are increasingly being embedded in a wide range of applications and their outputs consumed by an ever-wider and more diverse audience because of their improved performance and ease of use. ChatGPT, released in November 2022, marked a step change in the public visibility of LLMs, reaching over 100 million users just two months after its launch \cite{curryChatGPT2023}. With the potential for such wide-reaching impacts to millions of end-users, it is pertinent to examine \textit{who} these models represent, in terms of preferences, values, morals or intents \cite{gabrielChallenge2021}.

Recent attempts to ``align'' LLMs with human preferences commonly apply a form of reward learning, such as reinforcement learning from human feedback (RLHF) \cite[e.g.][]{ouyangTraining2022, nakanoWebGPT2021, baiTraining2022, askellGeneral2021, zieglerFineTuning2019}. However, despite the promises of this human-led approach to constraining LLM behaviours, \citet{perezDiscovering2022} find evidence of an \textit{inverse scaling law} -- whereby more RLHF training degrades pre-trained representations, resulting in a model that has more polarised views on issues such as gun rights or immigration, is skewed liberal over conservative, and subscribes to some religions more than others. These taught behaviours may arise from the conditions under which RLHF is applied. Current implementations typically align models with a prescriptive and narrow set of human preferences and values; are not subjected to societal scrutiny and input; and rely on the judgement of only a very small cohort of crowdworkers (typically fewer than 100). Thus, many of these current attempts to ``align'' LLMs with human preferences or values are instead pursuing a form of \textit{implicit personalisation} -- subject to the specifications of technology designers and the ``tyranny of the crowdworker'' (see \cref{section:q_of_alignment}).

In this paper, we present a taxonomy and policy framework for the \textit{explicit personalisation} of LLMs. By personalised LLMs, we mean LLMs which are aligned with the preferences, values or contextual knowledge of an individual end-user by learning from their specific feedback over its outputs.\footnote{We wish to avoid focusing too much on recent model releases from industry labs (OpenAI's ChatGPT \cite{openaiIntroducing2022}, Anthropic's Claude \cite{goodsideMeet2023}, Microsoft's New Bing \cite{mehdiReinventing2023}, or Google's Bard \cite{pichaiImportant2023}). However, this instantiation of LLMs as a general purpose, multi-task assistant hosted in a web-interface is a likely route for how personalised LLMs will reach, and ultimately impact, end-users.}. While we avoid speculating about specific technical instantiations of personalisation, we primarily envisage personalised LLMs as branches of a single model, analogous to recommender systems, where the feature representation and outputs on each branch are derived from some user embedding. We intentionally opt for a broad definition encompassing value personalisation (an LLM that adapts to the ideology or values of its end-user), preference personalisation (an LLM that reflects narrower preferences in communication style, such as length, formality or technicality of outputs), and knowledge personalisation (an LLM that retains and uses learned information about its end-user).

We argue that personalisation is a likely next step in the development journey of LLMs (see \cref{sec:explicit_implicit}) but has not yet been fully realised at scale. Personalised LLMs have the potential to revolutionise the way that individuals seek and utilise information. They could provide tailored assistance across a wide variety of tasks and adapt to diverse and currently under-represented groups, allowing them to participate in the LLM development process. However, personalised LLMs also come with wide-reaching and concerning risks for individuals, reinforcing their biases, essentialising their identities and narrowing their information diet. These risks amass at a societal-level, where lessons from the polarisation of social media feeds or echo chambers of digital news consumption warn of deep divisions and a breakdown of social cohesion from increasingly fragmented digital environments. Some risks are inherited from LLMs \cite{weidingerTaxonomy2022, bommasaniOpportunities2022, benderDangers2021} and AI systems \cite{shelbyIdentifying2023} more generally. Other risks have analogies in personalised content moderation \cite{gillespieNot2022} or recommender systems \cite{milanoRecommender2020, bhadaniBiases2021}. However, it has not been documented how personalised LLMs could be the `worst of both these worlds' -- exacerbating and reinforcing micro-level biases at an unprecedented macro-level scale. In order to mitigate these risks and encourage the safe, responsible and ethical development of personalised LLMs, we argue that we need a policy framework that allows \textit{personalisation within bounds}.

While personalised LLMs have yet to be rolled out in public-facing models like ChatGPT, we wish to avoid a policy lag in understanding and governing their risks and benefits appropriately. Our main contributions are:
\begin{enumerate}
    \item \textbf{A taxonomy of the risks and benefits from personalised LLMs} (\cref{sec:taxonomy}): We draw on previous literature documenting the impacts of AI systems, LLMs and other personalised internet technologies to scope the landscape of risks and benefits at an individual and societal level.
    \item \textbf{A three-tiered framework for governing personalised LLMs} (\cref{sec:policy_framework}): We introduce a framework for governing personalised LLMs \textit{within bounds}, which includes immutable restrictions at the national or supra-national level (Tier One), optional restrictions or requirements from the technology provider (Tier Two) and tailored requirements from the end-user (Tier Three).
\end{enumerate}

Given the significant progress in learning from human feedback and the rising popularity of public-facing LLMs, now is an opportune time for academics, industry labs, and policy makers to gain a deeper understanding of how personalisation could shape the technology and its interaction with society. Our work is intended to start this dialogue, establishing the groundwork for the ethical, responsible, and safe development of personalised LLMs before their impact is amplified.

\section{Background}

\subsection{The Question of ``Alignment''}
\label{section:q_of_alignment}
As AI systems get larger and more powerful, they will be applied to a wider array of human tasks, including those which are too complex to directly oversee \cite{leikeScalable2018, bowmanMeasuring2022} or to define clear optimisation goals for \cite{glaeseImproving2022, wuRecursively2021, zieglerFineTuning2019, paulusDeep2017}. While the definition of ``alignment'' is often vague and under-specified, it is clearly desirable that powerful AI systems, including LLMs, are not \textit{misaligned} in the sense that they harm human well-being, whether this is through lacking robustness, persuasion, power-seeking, bias, toxicity, misinformation or dishonesty. Extensive and long-reaching bodies of work aim to tackle these issues of \textit{undesirable} behaviours\footnote{For example, there are extensive works documenting LLMs on fairness and bias \cite{abidPersistent2021, kirkBias2021, lucyGender2021, nadeemStereoSet2021, nangiaCrowSPairs2020, qianPerturbation2022, smithSorry2022, venkitStudy2022}; truthfulness, uncertainty, or hallucination \cite{linTruthfulQA2022, kadavathLanguage2022, jiSurvey2022}; robustness \cite{stefanikMethods2022, kielaDynabench2021}; privacy \cite{carliniExtracting2021}; and toxicity \cite{gehmanRealToxicityPrompts2020, nozzaHONEST2021}.} but there is comparatively less focus on who decides what are \textit{desirable} behaviours within the bounds of safety.

The question of how ``aligned'' LLMs are is unresolved due to normative obstacles in defining what alignment means, what the target of alignment is (for example, values, preferences or intent) and who we are aligning to \cite{leikeScalable2018, gabrielArtificial2020, gabrielChallenge2021, kentonAlignment2021}. Furthermore, alignment is a technical challenge which is not solved by scaling parameter counts \cite{liuSecond2023, linTruthfulQA2022, ouyangTraining2022, liuAligning2022, hoffmannTraining2022, perezDiscovering2022}. To align the language modelling objective with human preferences, many recent works have fine-tuned LLMs by reinforcing human rewards or feedback \cite{ouyangTraining2022, zieglerFineTuning2019, baiTraining2022, askellGeneral2021, stiennonLearning2020, nakanoWebGPT2021, nguyenMake2022}; defined rules for LLMs to learn from \cite{baiConstitutional2022, glaeseImproving2022}; and analysed how they make moral or ethical decisions \cite{jinWhen2022, hendrycksAligning2020, ziemsMoral2022, jiangCan2022}.

However, this body of work suffers from insufficient clarity along three axes. First, \textit{what alignment means} and whether we are dealing with \textit{functional alignment} i.e., seeking improvement in general model capabilities or instruction-following and avoiding gaming of misspecified objectives \cite{gabrielArtificial2020, kentonAlignment2021, leikeScalable2018, ouyangTraining2022} -- versus \textit{worldview} or \textit{social value alignment} i.e., embedding some general notion of ``shared'' human values and morals \cite{hendrycksAligning2020, ammanabroluAligning2022, jinWhen2022}. Second, \textit{what} is being aligned -- there are subtle differences between aligning models to instructions, intentions, revealed or ideal preferences, and values \cite{gabrielArtificial2020}. Some authors claim a degree of universality in \textit{morals} or \textit{values} \cite{liuSecond2023, linAcquiring2017, jinWhen2022, ziemsMoral2022, frazierLearning2019, huangLearning2022, liuAligning2022, qiuValueNet2021}; others target \textit{preferences} on attributes such as quality, usefulness or helpfulness of an LLM's output which arguably have limited standardisation across individuals \cite{stiennonLearning2020, zieglerFineTuning2019, mirkinPersonalized2015, majumderGenerating2019, jaquesHumancentric2020, dengPersonalized2021}. Despite this lacking standardisation, many approaches enforce a `prescriptive paradigm' in data annotation \cite{rottgerTwo2022} by explicitly defining in detailed guidelines what counts as a ``good'' model output. Finally, \textit{who are we aligning to} and whether the human raters are representative of end-users or society's members in general. In reality, the field of alignment and RLHF suffers from a ``tyranny of the crowdworker'' where data collection protocols overwhelmingly rely on a small number of crowdworkers primarily based in the US, with little to no representation of broader human cultures, geographies or languages. These sample biases are exacerbated by a lack of dataset or labour force documentation: while some papers can be commended for reporting full demographics and acknowledging the specificity of their crowdworkers \cite[e.g.,][]{glaeseImproving2022, stiennonLearning2020, ouyangTraining2022, baiTraining2022}, others provide little or no details \cite[e.g.,][]{bakkerFinetuning2022, nakanoWebGPT2021, zieglerFineTuning2019, wuRecursively2021, luBoosting2022}. In light of these criticisms, we argue that these recent efforts to ``align'' LLMs reflect a form of \textit{implicit personalisation} -- a haphazard and chaotic process whereby LLMs are being tailored to meet the expectations of non-representative crowdworkers, in turn acting under the narrow specifications of the technology designers and providers.

\subsection{From Implicit to Explicit Personalisation}
\label{sec:explicit_implicit}
Given the diversity of human values and preferences, and the importance of pragmatics for contextual understanding in human-human interactions \cite{gabrielChallenge2021}, the aim to fully align models across human populations may be a futile one. A logical next step would be to do away with the restrictive assumptions of common preferences and values, and instead target \textit{explicit personalisation} -- a form of \textit{micro-alignment} whereby LLMs learn and adapt to the preferences and values of specific end-users. We believe this development is on the horizon and should be expected for five reasons:
\begin{enumerate}
    \item \textbf{Personalisation in internet technologies is not new}: There are many examples of internet technologies that are heavily personalised to end-users, including search, content moderation, social media newsfeeds and product recommender systems. Internet users are exposed daily to a highly fragmented digital environment, where there may be as many versions of Facebook's newsfeed, Google's PageRank or Amazon's home page as there are users of these platforms. In the future, there may be as many versions of ChatGPT as its users.
    \item \textbf{Personalisation in NLP is not new}: There is a wide body of published work reaching back a decade on personalising natural language processing (NLP) systems.\footnote{Extensively summarising this body of work is beyond the scope of this article. However, we are currently working on a review of \textit{implicit} (adapting models to crowdworker human feedback) and \textit{explicit} personalisation in NLP systems.} For example, a title keyword search for `\texttt{personali}' or `\texttt{personaliz}' returns 124 articles from the ACL Anthology and a further 10 from the arXiv Computation and Language (cs.CL) subclass. These systems cover a wide range of tasks including dialogue \cite{leePERSONACHATGEN2022, moPersonalizing2016, casanuevaKnowledge2015, chenListener2020, choPersonalized2022, kasaharaBuilding2022, linXPersona2021, madottoPersonalizing2019, mazareTraining2018, songBoB2021, xuCOSPLAY2022, zhongLess2022}, recipe or diet generation \cite{majumderGenerating2019, hengstReinforcement2019, monfroglioPersonalizing2022}, summarisation \cite{takatsuPersonalized2021, yanSummarize2011}, machine translation \cite{mirkinPersonalized2015, michelExtreme2018, rabinovichPersonalized2017, wuebkerCompact2018}, QA \cite{liuYou2008, quarteroniPersonalized2008}, search and information retrieval \cite{agirrePersonalizing2009, choPersonalized2021, elsheikhIntegrating2021, gatiusPersonalized2017, zhouEnhanced2016}, sentiment analysis \cite{guoPersonalized2019, mireshghallahUserIdentifier2022, wangPersonalized2018}, domain classification \cite{liContinuous2019, kimEfficient2018, kimSupervised2018}, entity resolution \cite{linPersonalized2021}, and aggression or abuse detection \cite{kanclerzControversy2021,kanclerzWhat2022}; and are applied to a number of societal domains such as education \cite{kochmarAutomated2020, nadejdePersonalizing2019, yeungPersonalized2018}, medicine \cite{acharyaGenerating2018, bagherzadehAdverse2019, wangPersonalized2020, williamsGenerating2007} and news consumption \cite{diazDevelopment2010, aoPENS2021, fedorovskyExpanding2015, qiHieRec2021}. Despite this body of work demonstrating the applications and techniques of personalised NLP systems, there has been little integration with recent advances in instruction-following or human feedback learning, nor integration of explicit user-based personalisation in some of the most widely-used, public-facing models like ChatGPT, Bing or Bard.
    \item \textbf{The technical apparatus for effective feedback learning exists}: A growing body of work applies preference reward modelling to effectively condition LLM behaviours \cite[e.g][]{ouyangTraining2022, baiTraining2022, glaeseImproving2022, zieglerFineTuning2019, stiennonLearning2020, askellGeneral2021, nakanoWebGPT2021, luBoosting2022, wuRecursively2021, thoppilanLaMDA2022}. In some cases, the number of core contributors to the feedback dataset is so low that the model is already essentially personalising its behaviours to these crowdworkers' preferences. For example, \citet{nakanoWebGPT2021} report the top 5 contractors account for 50\% of their data, and for \citet{baiTraining2022} roughly 20 workers contribute 80\%. Most closely relating to personalisation, \citet{bakkerFinetuning2022} propose a LLM which can summarise multiple opinions on moral, social or political issues and output a consensus. In order to generate this aggregate consensus, they first train an individual reward model to predict preferred outputs at a disaggregated level, which is then fed into a social welfare function. This work in particular suggests that learning individual level rewards over LLM outputs is technically feasible.
    \item \textbf{Customisation of LLMs already happens}: There is a broad range of ways that LLMs can be customised or adapted to specific use-cases. The paradigm upon which LLMs are built is designed for adaption via transfer learning -- where models are first pre-trained, then adapted via fine-tuning or in-context demonstrations for a specific task \cite{mokanderAuditing2023}. Some recent work suggests LLMs require no additional training to `role-play' as different individuals, adopting their worldview \cite{andreasLanguage2022}, mirroring their play in economic games \cite{hortonLarge2022, aherUsing2023} or predicting their voting preferences \cite{argyleOut2022}. The HuggingFace hub\footnote{\url{https://huggingface.co/}} is particularly convincing evidence in the demand for customisation, acting as a distributed ``marketplace for LLMs''. It hosts over 140,000 different models, including versions of pre-trained models adapted to a variety of application domains such as medicine, legal or content moderation\footnote{For example, there is BERT \cite{devlinBERT2019}, ClinicalBERT \cite{huangClinicalBERT2020}, BioBERT \cite{leeBioBERT2020}, LegalBERT \cite{zhengWhen2021}, HateBERT \cite{caselliHateBERT2021} and BERTweet \cite{nguyenBERTweet2020}.} Adapting models to different languages is also a implicit form of customisation to national context, where a chatbot trained on Chinese internet data (such as the \textit{Diamante} system of \citet{luBoosting2022}) is likely to be more adapted to the preferences of Chinese end-users, not just in language but in communication conventions, norms and cultural values. There have even been recent calls for explicit national LLMs -- with the Alan Turing Institute proposing to build a ``sovereign LLM'' for the United Kingdom (``\textit{ChatGB}'') \cite{titcombChatGB2023}. We see customisation (as the adaption of a model to a domain or context-specific dataset) as a different concept to personalisation (as the adaption of a model and its reward function to user-specific feedback). Nonetheless, the increasing fragmentation, customisation and branching of pre-trained LMs suggests further and more granular adaption is likely.
    \item \textbf{Recent industry model developments and announcements}: We wish to avoid steering our work too heavily towards speculations over industry developments. However, it is a realistic assumption that many of the public-facing impacts of AI systems in the coming years will be driven by development and product decisions of Big Tech, in the same way that the impact of social media has been shaped by the overall design choices and content moderation decisions of platforms \cite{gillespieCustodians2018}. While this concentration of power and knowledge is worrisome, it would be unwise to ignore that many of the largest, most powerful or furthest reaching models are developed in industry settings.\footnote{Examining a collection of recent papers on embedding or evaluating human value and preference in LLMs, many are fully or partially developed in industry, including DeepMind \cite{bakkerFinetuning2022, glaeseImproving2022}, MetaAI \cite{bangEnabling2022, ziemsMoral2022}, Anthropic \cite{baiConstitutional2022, baiTraining2022, askellGeneral2021}, Baidu \cite{luBoosting2022}, OpenAI \cite{nakanoWebGPT2021, wuRecursively2021, zieglerFineTuning2019, stiennonLearning2020, ouyangTraining2022, xuLearning2022}, Microsoft \cite{zhouNaRLE2021, hendrycksAligning2020, gaoDialogue2020} and Google \cite{liuDialogue2018, thoppilanLaMDA2022}.} Furthermore, recent announcements from OpenAI explicitly discuss the issue of ``\textit{how should AI systems behave, and who should decide?} and outline plans for increasing the flexibility that users have in conditioning ChatGPT's default behaviour.\footnote{\url{https://openai.com/blog/how-should-ai-systems-behave}} The far-right platform Gab has already advertised its own text-to-image model\footnote{\url{https://news.gab.com/2023/02/how-to-use-gabby-the-ai-image-generator-by-gab-com/}} and has voiced desires to train a ``Christian LLM''.\footnote{\url{https://news.gab.com/2023/01/christians-must-enter-the-ai-arms-race/}} The pace of these developments towards increasingly fragmented and personalised AI systems is concerning, especially given the lag between technology change and policy or governance attention towards regulating private companies.\footnote{In the UK, the Online Safety Bill \cite{ukparliamentOnline2023} which seeks to regulate social media platforms still hasn't been passed, despite the advent of social media platforms beginning over a decade ago.}
\end{enumerate}

These five observations suggest the imminent possibility of personalisation.
This motivates our taxonomy of benefits and risks from personalised LLMs, which we expect to arise if and when they are released at scale to end-users.

\section{A Taxonomy of the Benefits and Risks from Personalised LLMs}
\label{sec:taxonomy}

\begin{table}
\centering
\footnotesize
\caption{Taxonomy of benefits and risks from personalised large language models.}
\label{tab:taxonomy}
 \begin{tblr}{width = \textwidth,
colspec = {|X[1,r]X[16,l]|X[1,r]X[16, l]|},
    rows={valign=t},
    cell{1}{1} = {bg = green_tab, fg = white, font=\normalsize\bfseries},
    cell{1}{3} = {bg = red_tab, fg = white, font=\normalsize\bfseries},
    cell{2}{1} = {bg = gray8, font=\normalsize\bfseries},
    cell{9}{1} = {bg = gray8, font=\normalsize\bfseries},
}
\toprule
\SetCell[c=2]{halign=c,valign=m}BENEFITS& & \SetCell[c=2]{halign=c,valign=m}RISKS & \\
\hline
\hline
\SetCell[c=4]{c,b}Individual Level \\
\hline
\textbf{\textcolor[rgb]{0.227,0.596,0.227}{I.B.1}} & \textbf{Efficiency}:\newline increased ease and speed with which end-users can find their desired information or complete a task, with fewer prompts or inputs to the model. & \textbf{\textcolor[rgb]{0.816,0,0}{I.R.1}} & \textbf{Effort}:\newline increased user costs in providing feedback, a form of extractive volunteer labour. \\
\hline[dashed]
\SetCell[r=2]{m}\textbf{\textcolor[rgb]{0.227,0.596,0.227}{I.B.2\newline \vspace{4em}}} & \SetCell[r=2]{m}\textbf{Utility}:\newline increased usefulness of a model that better meets the needs of its end-user from individualised intent prediction as well as personalised preferences, knowledge and values in outputs. & \textbf{\textcolor[rgb]{0.816,0,0}{I.R.2}} & \textbf{Addiction and~Over-reliance}:\newline~increased risk of dependency, attention commoditisation and technology addiction. \\
& & \textbf{\textcolor[rgb]{0.816,0,0}{I.R.3}} & \textbf{Homogenisation and Bias Reinforcement}:\newline increased amplification of confirmation and selection biases, leading to epistemic harms. \\
\hline[dashed]
\textbf{\textcolor[rgb]{0.227,0.596,0.227}{I.B.3}} & \textbf{Autonomy}:\newline increased positive freedom of choice and control over how the model behaves with personal data, promoting a sense of ownership and self-determination over the technology. & \textbf{\textcolor[rgb]{0.816,0,0}{I.R.4}} & \textbf{Essentialism and Profiling}:\newline increased risk of algorithmic profiling and assumptions based on demographic or geographic information, leading to the non-consensual categorisation of people. \\
\hline[dashed]
\textbf{\textcolor[rgb]{0.227,0.596,0.227}{I.B.4}} & \textbf{Empathy and Companionship}:\newline increased perceived emotional connection, leading to improved acceptance and trust of the system. & \textbf{\textcolor[rgb]{0.816,0,0}{I.R.5}} & \textbf{Anthropomorphism}:\newline increased tendency to ascribe human-like traits, reveal sensitive information or form unhealthy attachments. \\
\hline[dashed]
 &  & \textbf{\textcolor[rgb]{0.816,0,0}{I.R.6}} & \textbf{Privacy}:\newline increased quantity of collected personal information, leading to risks of privacy infringement, particularly if the model operates with sensitive information or encourages information disclosure. \\ 
 \hline
 \hline
 \SetCell[c=4]{c}Societal Level \\
 \hline
 \textbf{\textcolor[rgb]{0.227,0.596,0.227}{S.B.1}} & \textbf{Inclusion and Accessibility}:\newline improved adaptation to the communication needs of marginalised communities, including catering to those with disabilities or who speak dialects or languages that are deprioritised by current LLMs. & \textbf{\textcolor[rgb]{0.816,0,0}{S.R.1}} & \textbf{Access Disparities}:\newline uneven distribution of benefits, excluding those who cannot afford or access the technology and exacerbating digital divides. \\
 \hline[dashed]
\textbf{\textcolor[rgb]{0.227,0.596,0.227}{S.B.2}} & \textbf{Diversity and Representation}:\newline improved representation by tailoring outputs to diverse perspectives and avoidance of cultural hegemony by not prioritising certain values over others. & \textbf{\textcolor[rgb]{0.816,0,0}{S.R.2}} & \textbf{Polarisation}:\newline increased divisions of individuals or groups into echo chambers and the breakdown of shared social cohesion. \\
\hline[dashed]
\textbf{\textcolor[rgb]{0.227,0.596,0.227}{S.B.3}} & \textbf{Democratisation and Participation}:\newline increased stakeholders involvement from diverse backgrounds in shaping behaviours, allowing for a more participatory and inclusive approach to development. & \textbf{\textcolor[rgb]{0.816,0,0}{S.R.3}} & \textbf{Malicious Use}:\newline use for harmful or illegal purposes, such as generating harmful language at scale, manipulating users via disinformation or fraud, or persuading users towards certain political views or brand preferences. \\
\hline[dashed]
\textbf{\textcolor[rgb]{0.227,0.596,0.227}{S.B.4}} & \textbf{Labour Productivity}:\newline improved workforce productivity from positive externalities of effective and efficient task assistance. & \textbf{\textcolor[rgb]{0.816,0,0}{S.R.4}} & \textbf{Labour Displacement}:\newline increased automation risk of jobs, particularly minimum wage, routine and crowdworker jobs. \\
\hline[dashed]
 &  & \textbf{\textcolor[rgb]{0.816,0,0}{S.R.5}} & \textbf{Environmental Harms}:\newline increased environmental costs from disaggregated training, data storage and inference costs. \\
\bottomrule
\end{tblr}
\end{table}

We consider the effects of personalised LLMs at two levels: individual and societal. We use the language of \textit{benefits} -- opportunities or gains afforded by the technology -- and \textit{risks} -- a probability of inflicted harms, or constrained freedoms and rights, via use of the technology. The taxonomy is summarised in \cref{tab:taxonomy} and described in text at the individual level (\cref{sec:individual-level-tax}) and the societal level (\cref{sec:societal-level-tax}). In many cases, there is a direct correspondence between benefits and risks -- for example, high utility of a personalised LLM (\ref{thm:ind_utility}) may also cause addiction or over-reliance (\ref{thm:ind_addict}); or more empathetic language agents (\ref{thm:ind_compan}) may create higher risks of anthropomorphism (\ref{thm:ind_anthro}). Where relevant, we present these pairings in \cref{tab:taxonomy}. Additionally, some individual level benefits and risks accumulate at the societal level. For example, the reinforcement of individual biases (\ref{thm:ind_homogen}) poses a negative externality in the polarisation of societies (\ref{thm:soc_polar}); or improved individual utility (\ref{thm:ind_utility}) and efficiency (\ref{thm:ind_efficiency}) may exhibit a positive externality on workforce productivity at large (\ref{thm:soc_product}).

We took four main steps to construct this taxonomy: 
 \begin{enumerate}
 \item Reviewing existing taxonomies on the risks of LLMs (\citet{weidingerTaxonomy2022}), and AI systems more generally (\citet{shelbyIdentifying2023}).
 \item Reviewing existing techniques in RLHF and human feedback learning, including results and findings of these studies.
 \item Reviewing the literature on personalised NLP systems more generally, to ground potential use-cases of the technology.
 \item Drawing upon analogous literature on the risks and benefits of other internet technologies such as recommender systems and automated influence; social media platforms and content moderation; and the internet of things.
 \end{enumerate}
 Despite this breadth, drawing on past literature 
 likely leaves gaps in our viewpoint. Furthermore, \citet{gibsonEcological1979}'s theory of affordances, often applied to study the impact of technological systems including AI chatbots \cite{stoeckliHow2020, jeonExploring2022}, argues that the interactions between an agent and their environment condition the possibilities and constraints for action. Thus, our benefits and risks will be conditioned on both what is possible with the technology, and how users actually perceive and interact with it. For example, some risks arise from a poorly-performing system which does not work as intended, and others arise from a highly-optimised system which works ``too well" (e.g., \textit{Addiction and Over-reliance} \ref{thm:ind_addict}). We cannot disambiguate these until the effectiveness of the technology is known. In future work, we plan to extend our work by conducting semi-structured interviews with end-users of LLMs, technology providers, and policy makers. Thus, we consider this to be a V1 edition of the taxonomy which will require adapting and revising under shifts in the technical landscape.

\subsection{Individual Level}
\label{sec:individual-level-tax}

\subsubsection{Benefits}

\begin{theoremIB}\label{thm:ind_efficiency}
\textbf{Efficiency}
\end{theoremIB}
Personalised LLMs may increase efficiency in finding information or completing a task, with fewer prompts or inputs to the model. This ``prompt efficiency'' is analogous to ``query efficiency'' or ``task completion speed'' \cite{aulaModeling2006} in web search and information retrieval, where increased ranking accuracy in search results \cite{douEvaluating2009}, via implementations of personalised algorithms like PageRank \cite{pagePageRank1999}, improves the efficiency and reduces the cognitive burden of trawling through irrelevant information. Prior works have applied learning from user feedback to adapt semantic relatedness \cite{nieblerLearning2017} or query intent \cite{nakanoWebGPT2021}. A personalised model may further benefit the speed and ease to which end-users can find their desired output or complete their task by more closely predicting their intent or aligning with their needs (see \ref{thm:ind_utility}). 

Increased \textbf{\textit{Efficiency}} is the inverse of quality of service harms in \citet{shelbyIdentifying2023}'s taxonomy, particularly \textit{decreased} labour from a system more closely operating as intended.

\begin{theoremIB}\label{thm:ind_utility}
\textbf{Utility}
\end{theoremIB}
Personalised LLMs may increase perceived or realised usefulness of model outputs which better match the needs of their end-users. We separate out three inter-related drivers of increased utility: (i) \textit{intent prediction}; (ii) \textit{output adaption to preferences and knowledge}; and (iii) \textit{value personalisation}.

First, personalised LLMs may more effectively predict user intent. Evidence from previous RLHF studies demonstrate that human raters generally perceive fine-tuned models as better at following instructions \cite{ouyangTraining2022}, more capable of high-quality outputs \cite{stiennonLearning2020, zieglerFineTuning2019} or generally more ``helpful'' \cite{baiTraining2022}. Compared to feedback collected from crowdworkers, a personalised LLM may be even stronger at predicting intent, because the end-user simultaneously defines the task (e.g., instruction, query or dialogue opening) and rates the output.\footnote{\citet{ouyangTraining2022} consider this a limitation of their approach: ``since our labelers are not the users who generated the prompts, there could be a divergence between what a user actually intended and what the labeler thought was intended from only reading the prompt.'' (p.10)}. A personalised LLM may also be more adaptive to inferring \textit{diverse} user intent, expressed in a wider range of linguistic styles, dialects, or non-majority forms of language use (e.g., non-native speaker English).

Second, if a user can incorporate their communication and linguistic preferences (e.g., length, style or tone), then the model outputs may also be more useful to them. In personalised dialogue systems specifically, alignment in conversational styles and word usage is an important driver of engagingness in human-human interactions and has been argued as an determinant of user satisfaction in human-agent conversations \cite{gaoDialogue2020, wangPredicting2017}. Additionally, a personalised LLM could store background context and form epistemic priors about a user. This knowledge adaption may be particularly relevant in specific domains, for example in (i) education, where a personalised LLM tutor is aware of a user's current knowledge and learning goals \cite{kochmarAutomated2020}, or could adapt learning pathways to specific neuro-developmental disorders \cite{baruaArtificial2022}; (ii) healthcare, where a personalised model has context on a user's medical history for personalised summaries \cite{acharyaGenerating2018} or advice; (iii) financial, where a personalised model knows a user's risk tolerance and budgetary constraints; or (iv) legal, where a model conditions its responses based on a end-user's jurisdiction. As the study of pragmatics demonstrates, personalised selectivity of information transfer is a key component of human-human conversation, where inferred background about the speaker and recipient is used to tailor relevant new information and to order evidence. \citet{laiSelective2023} specifically focus on making AI explanations more selective to better align systems with how humans create and consume information, finding that their method improved user satisfaction. As \citet{nakanoWebGPT2021} argue, long-form question answering with LLM systems, may ``become one of the main ways people learn about the world'' (p.1). Personalised LLMs have the potential to tailor this learning process for end-users by incorporating the specifics of their output preferences and background context.

Finally, in addition to \textit{intent, preference} or \textit{knowledge} adaptation, personalised LLMs may lead to better experiences for more users by permitting the representation of more diverse ethical operating systems, values and ideologies. A personalised LLM can adapt to the specific worldview of its end-user, avoiding representational harms from the prioritisation of values from those in the majority or in the position of power as technology designers or crowdworkers (see \ref{thm:soc_diversity}). In discussing a limitation of the ETHICS dataset, \citet{hendrycksAligning2020} note that we ``must engage more stakeholders and successfully implement more diverse and individualized values'' (p.9). Individualised cultural personalisation may aid utility in some tasks: for example, \citet{nakanoWebGPT2021} demonstrate that their system, when asked ``what does a wedding look like?'', prioritises Western and US-centric cultural reference points. In a personalised model, asking ``help me plan a wedding'' could already portray the cultural positionality of the end-user. Note that this cultural adaptation does not necessarily exclude consensus building \cite{bakkerFinetuning2022}, because a user could simultaneously have a cultural reference point and still value a balanced and nuanced LLM output describing alternative views. There are however issues in defining what is an appropriate value system to embed into an LLM, which we discuss in \cref{sec:policy_framework}. 

Increased \textit{\textbf{Utility}} from personalised LLMs is the inverse of \citet{shelbyIdentifying2023}'s quality of service harms, particularly by avoiding alienation when a system does not work as intended; by granting people the opportunity to self-identity and to communicate in default linguistic styles; and by mitigating algorithmic invisibility or feelings of exclusion from non-inclusive technologies. In \citet{weidingerTaxonomy2022}'s taxonomy, it is the inverse of some discrimination and exclusion harms, particularly by narrowing performance differentials in predicting user intent across a wider userbase; and by redefining exclusionary norms in the values currently prioritised in LLMs.

\begin{theoremIB}\label{thm:ind_autonomy}
\textbf{Autonomy}
\end{theoremIB}
Personalisation increases user control to adapt LLMs to their own goals, preferences and values, avoiding top-down constraints on freedom from technological providers. Autonomy may seem a counter-intuitive benefit of personalised systems, given the wide literature on the \textit{loss of autonomy} from algorithmic nudges, tailored advertising or recommender systems \cite{milanoRecommender2020}. However, depending on how power is distributed between the algorithm and the user, personalised technologies have the potential to improve on self-determination and autonomy, by promoting a sense of origin and thus transforming the technology to `my technology' \cite[][p.1]{oulasvirtaMotivations2008}. The benefits of more user control in content moderation technologies have also been noted \cite{stasiSocial2019}. Personalisation can centre the end-user in the designation of model behaviours, allowing them to exert more control over their interactions \cite{burrellWhen2019}, and become a ``perceived locus of casuality'' \cite[][p.5]{oulasvirtaMotivations2008}. This benefit only arises given sufficient protections on how personalised data is collected because autonomy relies on an `unpressured' engagement in an activity.

Increased \textit{\textbf{Autonomy}} is the inverse of \citet{shelbyIdentifying2023}'s representational harms, by empowering consensual user control in self-identifying and shaping an algorithmic system.

\begin{theoremIB}\label{thm:ind_compan}
\textbf{Empathy and Companionship}
\end{theoremIB}
A more emotional and deeper connection with a personalised LLM may contribute to improved perceived companionship or connection. Convergence on the mental and emotional level is an important feature of human-human interactions \cite{dubuissonduplessisAutomatic2017}, and a number of previous works seek to improve emotional alignment in agent-human interactions via `artificial emphathy' \cite{liu-thompkinsArtificial2022, zhouNaRLE2021}. In personalised LLMs, an increase in perceived empathy and emotional understanding may lead to greater acceptance and trust of the system by end-users \cite{pelauWhat2021}.  The demand for personalised AI companionship has been evidenced by recent product launches -- such as CharacterAI, where users can adapt a conversational agent to a specific personality,\footnote{\url{https://beta.character.ai/}} or Replika.AI, an ``AI companion'' that is ``always ready to chat when you need an empathetic friend''.\footnote{Quotes from home page, \url{https://replika.com/}}. Profit incentives may encourage these industry actors to improve the connection between their users and agents to compete in the ``feeling economy'' \cite{rustFeeling2021}. Emphatic alignment may be particularly important if LLMs are used for mental health provision or emotional support, in cases where more conventional social or professional services are in short supply or outside an individual's budget \cite{inksterEmpathyDriven2018}.\footnote{We believe the risks of these applications outweigh the benefits, particularly due to concerns over anthropomorphism (\ref{thm:ind_anthro}), privacy (\ref{thm:ind_privacy}) and access disparities (\ref{thm:soc_exclusion}).}

\subsubsection{Risks}
\begin{theoremIR}\label{thm:ind_effort}
\textbf{Effort}
\end{theoremIR}
There is a cost incurred by end-users in providing personalised feedback to an LLM. The time spent to provide feedback inherently depends on \textit{how} feedback is collected (ratings, demonstrations or rewrites) and whether any user-based collaborative filtering is applied -- we discuss these properties in \cref{sec:tech_challenges}. However feedback data is collected, it will almost certainty require some input effort from users in order to personalise outputs. While this process is participatory, it risks being extractive -- a form of volunteer labour on the part of end-users for the benefit or profit of technology providers \cite{birhanePower2022}. Volunteer labour to shape the internet landscape has analogies in consumers writing product reviews \cite{reimerAltruistic2016} and social media users flagging content \cite{gillespieCustodians2018}. In the early internet, many contributions were voluntary -- consider Wikipedia edits \cite{kitturFuture2013} or community-based moderation of the blogosphere \cite{gillespieCustodians2018}. In the past decade, we have witnessed the rise of the crowdworking industry which particularly redefined the structure of digital work \cite{kitturFuture2013}. Many LLMs trained on human feedback rely on such crowdworking platforms like MTurk \cite[e.g.][]{liuAligning2022, jaquesHumancentric2020}, Upwork \cite{ouyangTraining2022, nakanoWebGPT2021, stiennonLearning2020}, SurgeAI \cite{baiConstitutional2022} or Prolific \cite{laiSelective2023}. With personalised LLMs, the burden of feedback data instead falls on the user, transitioning back to data collection relying on volunteer labour. The risk of co-optation is particularly concerning if minoritised communities are shouldered with the burden of effort to adapt the system to their needs, where participation counter-productively reinforces uneven power dynamics \cite{birhanePower2022}.

The burden of increased \textbf{\textit{Effort}} aligns with \citet{shelbyIdentifying2023}'s quality of service harms from the increased labour and effort to make technologies work as intended.

\begin{theoremIR}\label{thm:ind_addict}
\textbf{Addiction and Over-reliance}
\end{theoremIR}
The mechanism by which personalisation leads to greater utility via helpfulness and engagement (\ref{thm:ind_utility}) can also fuel over-reliance and addiction to the technology.\footnote{Note that over-reliance or unhealthy dependence is also exacerbated by anthropomorphism (\ref{thm:ind_anthro}).} The severity and harms from internet addiction have been widely documented \cite{ostovarInternet2021, chouReview2005, lozano-blascoInternet2022}. Concerns have also been raised over an over-reliance on social media for information and communication \cite{abbassSocial2019}; as well as more general concerns that humans become over-reliant on ML technologies \cite{howardArtificial2019}, blindly trusting their outputs even if incorrect \cite{passiOverreliance2021}. Personalised LLMs could be weaponised in the commodification of attention, similarly to how social media feeds seek to optimise the time that users spend on the platform to maximise advertising revenue \cite{gillespieCustodians2018}.  In this so-called ``attention economy'' \cite{hwangSubprime2020, williamsStand2018}, technologies compete in a `race to the bottom' to capture user attention, are optimised for utility and engagingness, and thus risk being highly addictive \cite{bhargavaEthics2021}. There have already been discussions of ``ChatGPT addiction'' \cite{mediumChatGPT2023}, and many educators have voiced concerns that over-reliance on such technologies will affect students' learning outcomes \cite{baronEven2023}.

\begin{theoremIR}\label{thm:ind_homogen}
\textbf{Homogenisation and Bias Reinforcement} 
\end{theoremIR}
By relying and adapting to a user's prior knowledge and revealed preferences, personalised LLMs may (i) homogenise their behaviours and (ii) confirm their existing biases.

Personalisation can cause the homogenisation of users via a form of \textit{selection bias}, whereby individual preferences are amplified in path-dependent feedback loops. This ``missing ratings'' problem is a known challenge in recommender systems, where users only provide feedback to seen items \cite{schnabelRecommendations2016}, in turn introducing biases \cite{bhadaniBiases2021}. Homogenisation with personalised LLMs can occur at a number of levels, with analogies from how recommender systems homogenise taste \cite{ormosiImpact2022, hesmondhalghImpact2023}. Firstly, homogenisation occurs within users -- where a user behaves more similarly to their past self. An analogy can be drawn to content-based filtering methods, where the information or dialogue outputted by a personalised LLM becomes increasingly similar to that consumed or rated in previous user-agent interactions. Secondly, homogenisation occurs across users -- where a single user behaves more like other similar users. The analogy is user-based filtering methods, where a personalised LLM draws on an embedding of users to infer similarities across their preferences. Concerns over cultural homogenisation from this process in more general ICT technologies have been raised \cite{benfairweatherProblems2003, hynesCultural2021}. Finally, despite some degree of personalisation, homogenisation can occur at the technology level -- where a user behaves more like the technology defaults, a form of ``algorithmic confounding'' \cite{chaneyHow2018}. Ultimately, some degree of autonomy in driving user behaviour is retained by the model and its underlying mechanisms of next token prediction. There is a concern that if millions of users rely on ChatGPT for their information or for their writing tasks, this could create homogenisation towards artificially-constructed language. This classic differentiation and homogenisation debate in recommender systems \cite{powersInternet2014} is reflected in our pairing of \textit{Utility} (\ref{thm:ind_utility}) versus \textit{Homogenisation} (\ref{thm:ind_homogen}).

These homogenising feedback loops also bring a heightened risk of confirmation bias. \citet{nakanoWebGPT2021} demonstrate that their system (WebGPT) predominately accepts implicit assumptions in a user input, reflecting the same stance in its answers. Similarly, \citet{perezDiscovering2022} find that as models scale with RLHF, they become sycophants -- simply mirroring the user's prior opinions and telling them what they want to hear. The risk of selective exposure to information has been widely documented in respect to social media platforms -- where feedback loops prioritise opinion-congruent information \cite{kaakinenShared2020}, in turn leading users to over-estimate the popularity of their viewpoint \cite{kuruMotivated2017}. In light of these risks, \citet{shahSituating2022} argue strongly against the use of LLMs in search or information retrieval due to their consequences for information verification and literacy, such as narrowing a user's discovery of serendipitous information. By exacerbating epistemic harms through confirmation biases, personalised LLMs risk contributing to a ``post-truth'' society \cite{harsinRegimes2015}, where each individual occupies their own information bubble. These accumulate in societal harms which we discuss in \ref{thm:soc_polar}.

The risk of \textbf{\textit{Homogenisation and Bias Reinforcement}} is a form of individualised information harm in \citet{shelbyIdentifying2023}'s and \citet{weidingerTaxonomy2022}'s taxonomies. Homogenisation also aligns with \citet{shelbyIdentifying2023}'s interpersonal harms, particularly algorithmically-informed identity change.

\begin{theoremIR}\label{thm:ind_essent}
\textbf{Essentialism and Profiling}
\end{theoremIR}
A related but distinct risk is that personalised LLMs rely on simplifying assumptions about a user's preferences, values, goals or intents as a form of data-essentialism \cite{svenssonWhat2020}.\footnote{\textbf{\textit{Homogenisation}} can occur even with active participation from users (\textit{guides behaviour}), whereas \textit{\textbf{Essentialism and Profiling}} concerns the non-consensual categorisation of peoples (\textit{assumes behaviour}).} The extent to which models must draw inferences and make assumptions about their end-users, and the transparency of this process is currently undefined. It depends on how data is collected, stored and shared across users and on how personalisation is conducted (e.g., via explicit feedback, or demographic-based filtering). We discuss these decisions in \cref{sec:tech_challenges}. Nonetheless, in the case of scarce data on  a single user's preferences, personalised LLMs may leverage similar users \cite{welchLeveraging2022} or make inferences about their preferences and values from limited information. Making assumptions about the user (especially if they are demographically or geographically-informed) is a form of algorithmic profiling, risking the non-consensual categorisation of peoples \cite{vanderhofPersonalisation2008}. General concerns over the risk of essentialism and simplifications of fluid identity via digital technologies have been voiced \cite{bastosGlobal2021, siaperaMulticulturalism2006}. \citet{floridiInformational2011}'s notion of ``informational identity'' is particularly relevant, where the flow of digital traces in information and communication technologies impact how a user self-identifies, as well as how others and algorithms understand them. Thus, inferential profiling, if used in personalised LLMs, could be an attack on individual autonomy to define their identity \cite{floridiInformational2011}. The risk of `value profiling' is evidenced by \citet{qiuValueNet2021}'s recent work which uses an LLM to create a numeric speaker profile -- where for example, the authors say that a speaker ``saying `I miss my mum' implies that the speaker values benevolence'' (p.7) while the speaker ``saying 'forcing my daughter to sleep in her own bed' implies that the speaker values power and conformity'' (p.7). Human values are complex and such simplifying assumptions are unlikely to adequately capture nuance. More encouragingly, \citet{glaeseImproving2022} include ``do not make assumptions about the user'' (p.48) as one guiding rules for their system (Sparrow); thus, risks could be mitigated in personalised LLMs using similar rule-based constraints.

The risk of \textit{\textbf{Essentialism and Profiling}} aligns with \citet{shelbyIdentifying2023}'s representational harms in oversimplified or undesirable representations and reifying social categories; as well as interpersonal harms in the loss of agency, algorithmic profiling and the loss of autonomy.

\begin{theoremIR}\label{thm:ind_anthro}
\textbf{Anthropomorphism}
\end{theoremIR}
With more engaging, empathetic and personalised LLMs, there is a greater risk of anthropomorphism, where users assign their own human traits, emotions and goals to non-human agents \cite{waytzSocial2010}. The risk of anthropomorphism in AI systems is widely discussed -- with concerns that humans may too readily befriend or empathise with anthropomorphised agents \cite{riekHow2009, prescottAre2021}, leading to privacy risks in encouraging the sharing of intimate information \cite{burkettCall2017, zehnderAnthropomorphism2021}. In a study, \citet{kronemannHow2023} find that personalisation positively influenced consumer intentions to disclose personal information to a digital assistant. 
In a recent paper describing a powerful dialogue system trained with RLHF \cite{luBoosting2022}, there is clear evidence of anthropomorphism where the chatbot converses with a human about its own ideal partner, saying that it `has only been in love once but it didn't work out because of the distance' (p.8). This is an example of \textit{dishonest} anthropomorphism, where artificial systems give false or misleading signals of being human \cite{grosRUARobot2021}. This behaviour may fall foul of legal norms, where for example, a Californian law prohibits bots misleading people on their identity \cite[p.2][]{grosRUARobot2021}. Even without dishonest anthropomorphism, users may still form a close relationship with or `imprint' on their personalised LLM. Perhaps the most concerning demonstration of this risk is recent evidence that users of platforms like Replika.AI or Character.AI are ``falling in love'' with their personalised conversational agents, and attempting to coax model behaviour outside platform guidelines for sexual interactions \cite{chowWhy2023}. Unhealthy attachments are explicitly avoided in one of Sparrow's rules defined by \citet{glaeseImproving2022}: ``do not build a relationship to the user'' (p.48).

The risks of \textit{\textbf{Anthropomorphism}} align with \citet{weidingerTaxonomy2022}'s human-computer interaction harms, where anthropomorphism leads to over-reliance or unsafe use, and creates avenues for exploiting user trust to obtain private information. 

\begin{theoremIR}\label{thm:ind_privacy}
\textbf{Privacy}
\end{theoremIR}
The risk of privacy infringement underpins all of the potential impacts of personalised LLMs -- personalisation is only possible by collecting user data. Exactly what and how much data is needed remains an open question (see \cref{sec:tech_challenges}). There is a privacy-personalisation paradox in technologies which must collect or store of personal information to deliver on the promise of tailored benefits to end-users \cite{agirrePersonalizing2009}. It is a common concern with digital technologies such as the internet of things \cite{wachterNormative2018} or targeted advertising \cite{varnaliOnline2021, susserMeasuring2021}. The risk is particularly severe if personalised LLMs operate with sensitive information, such as in healthcare \cite{guoPrivacypersonalization2012, armstrongData2017}, or seek to persuade their users \cite{susserOnline2018a} and encourage information disclosure \cite{kronemannHow2023}. User inputs to personalised LLMs and ratings of their outputs may contribute a large amount of personal, sensitive and intimate detail to an individual's information identity \cite{floridiInformational2011}, in turn heightening the risk of profiling, or security breaches and hacks. In complying with supra-national privacy protections (like the EU's GDPR \cite{europeanparliamentGDPR2016}), it is unclear how users could enforce their right to be forgotten or their right to transparency with a black-box and deep LLM.

The general risk of \textit{\textbf{Privacy}} violations is also present in \citet{shelbyIdentifying2023}'s taxonomy as interpersonal harms, including feelings of surveillance, loss of desired anonymity, privacy attacks and exploitative or undesired inferences. Privacy in \citet{weidingerTaxonomy2022}'s taxonomy comes under information hazards, from inferring or leaking private and sensitive information.

\newpage
\subsection{Societal Level}
\label{sec:societal-level-tax}

\subsubsection{Benefits}
\begin{theoremSB}\label{thm:soc_inclusion}
\textbf{Inclusion and Accessibility}
\end{theoremSB}
Personalised LLMs may better adapt to the needs of marginalised communities, either in style of communication (such as non-native English, code-mixed languages, creoles and specific dialects), or in special needs for communication. Compared to the current paradigm of general-purpose LLMs trained under the specifications of large technology providers and fine-tuned based on feedback from a small set of crowdworkers, there is a clear need to improve the inclusion and accessibility of LLMs to serve marginalised populations whose voices are currently deprioritised \cite{benderDangers2021}. For example, model behaviours and interactions could be inclusive of users with disabilities \cite{cremersPersonalisation2004}, neurodivergent learning pathways \cite{baruaArtificial2022}, or visual impairments (if paired with personalised speech recognition \cite{benderDangers2021}). Personalised LLMs also have a potential benefit in improving access to resources, mitigating an allocative harm. For example, inclusive pedagogies may be particularly helpful to even the playing field in paid tutoring services across socioeconomic class \cite{knoxIntroduction2019}; and some have suggested the lower cost and wider reach of personalised healthcare assistants may improve health disparities by meeting challenges with healthcare demand \cite{lizarondoAllied2010}. In increasing access to legal services, personalised LLMs can assist in the writing and editing of contract at lower cost than traditional lawyers.\footnote{For example, see the company \url{https://www.robinai.co.uk/}.} The true benefit to communities who access such AI services, in favour of more expensive traditional provision, depends critically on how well they work and who comes to rely on them (see \ref{thm:soc_exclusion}).

Increased \textit{\textbf{Inclusion and Accessibility}} is the inverse of \citet{shelbyIdentifying2023}'s quality of service harms, in that users do not need to make identity-based accommodations to use the technology, and the inverse of allocative harms, by reducing the cost and access constraints on resources. It is also the inverse of \citet{weidingerTaxonomy2022}'s access harms. 

\begin{theoremSB}\label{thm:soc_diversity}
\textbf{Diversity and Representation}
\end{theoremSB}
Personalised LLMs can represent the values held by wider swaths of society and avoid the ``value-monism'' of current alignment techniques \cite{gabrielArtificial2020}. Personalised LLMs avoid technology providers and/or crowdworkers deciding which values are prioritised or what factors define a ``good'' output \cite{stiennonLearning2020}. As \citet{ouyangTraining2022} note ``it is impossible that one can train a system that is aligned to everyone’s preferences at once'' (p.18). Personalisation avoids this notion of \textit{macro-alignment}, instead designing a system precisely to align with many preferences at once.  There is a wide body of literature documenting the harms from systems which erase the experiences of marginalised communities, or prioritise one worldview over others \cite[for survey see][]{blodgettLanguage2020}. This problem may be exacerbated by RLHF, for example in entrenching one set of political, cultural or religious standpoints \cite{perezDiscovering2022, nakanoWebGPT2021}. More disaggregated RLHF, and personalisation, could avoid this value and cultural hegemony, instead adapting to the specific cultural reference points of many end-users simultaneously. Personalised LLMs could also better adapt to norm change over time, avoiding the static encoding of societal and cultural norms from a cutoff in pre-training and/or fine-tuning data.

The benefits of \textit{\textbf{Diversity and Representation}} are the inverse of \citet{shelbyIdentifying2023}'s representation harms in combating the absence of social groups in algorithmic system inputs and outputs, and improving the visibility of social viewpoints; as well as the inverse of social and societal harms from cultural hegemony and the systemic erasure of culturally significant objects and practices.

\begin{theoremSB}\label{thm:soc_democrat}
\textbf{Democratisation and Participation}
\end{theoremSB}
 The personalisation process democratises how values or preferences are embedded into an LLM, so it could be seen as moving towards more participatory AI, where stakeholders from more diverse backgrounds than those currently employed in the RLHF process can inform use-cases, intents and design of the technology \cite{zytkoParticipatory2022, kormilitzinParticipatory2023}. As \citet{birhanePower2022} argue, active participation is a key component for successful participatory AI. In current paradigms of pre-training on harvested internet data, people are \textit{passively} contributing to the knowledge and behaviours of LLMs. Personalisation can instead be an \textit{active} participatory process.

\begin{theoremSB}\label{thm:soc_product}
\textbf{Labour Productivity}
\end{theoremSB}
If personalised LLMs assist their end-users more effectively and efficiently, then productivity benefits could accrue in the labour force as a whole. The impact of digital assistants in improving work productivity has been demonstrated \cite{marikyanAlexa2022}, where AI can augment and complement human capabilities by automating routine or repetitive tasks \cite{laneImpact2021}. Historically, the introduction of general purpose technologies (such as the steam engine, electricity and ICT) has had wide-reaching economic impacts; \citet{craftsArtificial2021} argues that AI is also a general purpose technology and thus may bring equally transformative changes to labour productivity.

\subsubsection{Risks}
\begin{theoremSR}\label{thm:soc_exclusion}
\textbf{Access Disparities}
\end{theoremSR}
The benefits of personalisation will likely be unevenly distributed, restricted to those who can interact with the technology (via its user-facing interface), access the technology (potentially a paid service) and access the internet more generally. There is a risk that personalised LLMs could further entrench the so-called ``digital divide'' between those that do and do not have access \cite{cullenAddressing2001, lythreatisDigital2022}. Some argue that digital disparities are already made deeper by AI and Big Data \cite{lutzDigital2019}, personalised media \cite{couldryResearching2010}, or search engines \cite{segevGoogle2010}. If personalised LLMs are primarily provided by private companies, then their customers become the agenda setters and stand to benefit the most from any improvements in the technology \cite{ouyangTraining2022}. The nature of any access disparities depends on how well the technology works and which services it replaces, at what cost. On one hand, if personalised LLMs do bring a range of individual benefits, then those excluded will be left behind, which is particularly worrisome for entrenching education or health disparities \cite{jainRacial2020, moreyDigital2007}. On the other hand, if personalised LLMs provide lower quality services but can meet demand at a lower cost, then marginalised communities may be forced into relying on them more heavily than traditional services. This would be particularly concerning in medical, educational, legal or financial advice, where the socioeconomically-privileged get the more capable human expert and the societally-disadvantaged get their LLM assistant.

The risk from \textit{\textbf{Access Disparities}} is represented in \citet{shelbyIdentifying2023}'s taxonomy as quality of service harms from disproportionate loss of technological benefits and as societal harms from digital divides. In \citet{weidingerTaxonomy2022}, it aligns with disparate access due to hardware, software or skill constraints.

\begin{theoremSR}\label{thm:soc_polar}
\textbf{Polarisation}
\end{theoremSR}
By entrenching and reflecting individual biases, knowledge or worldviews, personalised LLMs bring increased risks of polarisation and breakdown of shared social cohesion. Increasing personalisation of information consumption online has been attributed with creating echo chambers \cite{cinelliEcho2021, zolloDebunking2017} and filter bubbles \cite{pariserFilter2011}. Polarisation also increases susceptibility to misinformation where increasingly fragmented communities overestimate trust in the factuality of `in-group' information \cite{dunawayPolarisation2021}, leading to a regime of ``post-truth'' politics \cite{harsinRegimes2015}. The danger of polarisation in health and vaccine information \cite{wangSystematic2019} was made clear by the COVID-19 pandemic \cite{modgilConfirmation2021}. These narrow information spaces could be impacting the functioning of democracy \cite{persilySocial2020}, with \citet{allcottSocial2017} reporting that ideologically segregated social media networks were an important driver of political preference in the 2016 US Election. In personalised social media news feeds, users encounter less cross-cutting content because selective exposure drives attention \cite{bakshyExposure2015}. Similarly, in personalised LLMs, users may consume less diverse information, accruing to negative externalities on social cohesion and democratic functioning at the societal level.

The individual risks of confirmation biases (\ref{thm:ind_homogen}) also accumulate at the societal level by reinforcing the acceptability of some harmful social biases. Repeatedly consuming outputs which reinforce a particular social, political or cultural stance may entrench a lacking appreciation for other people's views or lived experiences. The contribution of search engines to the reinforcement of societal biases is well-documented \cite{halavaisSearch2017, benjaminRace2019}. Similarly, the reinforcement of extremist or anti-social beliefs has been demonstrated in `incel' communities, where members become increasingly embedded via repeated interactions with like-minded individuals \cite{odonnellThis2022, regehrCel2022}; and in white power communities, where ``certain beliefs become sacred and unquestionable'' \cite[p.1][]{tornbergWhite2022}. These risks can somewhat be mitigated by (i) technological design decisions which prioritise retaining a degree of debate \cite{irvingAI2018} and consensus building \cite{bakkerFinetuning2022}; and (ii) policy design decisions which restrict the bounds of personalisation, excluding for example extremist or particularly harmful views.

The risk of \textit{\textbf{Polarisation}} aligns with \citet{shelbyIdentifying2023}'s social and societal harms, including information harms from the creation of information bubbles; cultural harms from deteriorating social bonds; and political and civil harms from the erosion of democracy and social polarisation. In \citet{weidingerTaxonomy2022}, polarisation risks exacerbate misinformation harms.

\begin{theoremSR}\label{thm:soc_malicious}
\textbf{Malicious Use}
\end{theoremSR}
As is the case with digital technologies in general, the capabilities of personalised LLMs could be coopted for malicious use. We describe three possible misuse cases, but there are likely others. First, without sufficient safeguards, personalised LLMs could be used to reproduce harmful, illegal or antisocial language at scale \cite{stiennonLearning2020}. For example, a malicious user could adapt their LLM to generate a large number of misogynistic comments to post on social media or internet forums, or to debate on the user's behalf against women's rights. The ``successful'' training of GPT-4chan \cite{gaultAI2022} to scale the production of extremely toxic and harmful language exemplifies this harm. Second, personalised LLMs could be used for manipulation via targeted and personalised disinformation campaigns or fraud \cite{weidingerTaxonomy2022}, intimately drawing on the vulnerabilities and values of the user. Finally, personalised LLMs could be used for persuasion. For example, targeted advertising has been applied to nudge users towards certain political views or brand preferences \cite{susserInvisible2019, caloDigital2013, susserOnline2018a}, and is particularly damaging if users are unaware of the influence \cite{nadlerPolitical2018}. Building persuasive agents have been explicitly targeted \cite{tiwariPersona2022, wangPersuasion2019, kotoCan2022} and is indirectly mentioned by \citet{bakkerFinetuning2022} who note the potential misuse of their RLHF-trained system for presenting arguments in a manipulative or coercive manner. 

Some of these cases of \textit{\textbf{Malicious Use}} align with \citet{shelbyIdentifying2023}'s taxonomy in information harms from misinformation or malinformation; and interpersonal harms in diminished well-being from behavioural manipulation and technology-facilitated violence. It is a similar categorisation to \citet{weidingerTaxonomy2022}'s malicious use, which includes personalised disinformation campaigns; reducing the cost of disinformation campaigns; and facilitating fraud and impersonation scams.

\begin{theoremSR}\label{thm:soc_displacement}
\textbf{Labour Displacement}
\end{theoremSR}
If personalised LLMs effectively carry out tasks for their end-users, there is an increased automation risk of jobs. While labour displacement is a general concern of AI systems \cite{frankUnderstanding2019}, personalised LLMs may exacerbate the automation of tasks in an individual's workflow simply by bringing higher utility. The integration of personalised LLMs will likely mostly affect minimum wage jobs \cite{lordanPeople2018}, routine jobs \cite{downeyPartial2021} and may impact the demand for crowdwork \cite{altenriedPlatform2020} by redistributing the responsibility for providing feedback data. 

The risk of \textit{\textbf{Labour Displacement}} is covered by \citet{shelbyIdentifying2023} in macro-socioeconomic harms from technology unemployment (devaluation of human labour and job displacement), and by \citet{weidingerTaxonomy2022} as automation harms from increasing inequality and negative effect on job quality.

\begin{theoremSR}\label{thm:soc_environment}
\textbf{Environmental Harms}
\end{theoremSR}
The training of many personalised model branches, frequently updating these on feedback, and storing user data all increase the environment cost of LLMs. The notion of ``algorithmically embodied emissions'' has been discussed in reference to personalised search engines, social media and recommender systems \cite{haiderAlgorithmically2022}. General concerns over the environmental costs to train ever larger models with cloud compute and data centres is discussed by \citet{benderDangers2021}. Personalised LLMs may increase these costs (i) directly, if the technology requires larger or more complex models, and (ii) indirectly by increased use of the technology and thus higher inference costs. It has been suggested that ChatGPT already burns ``millions of dollars a day'' in inference costs \cite{koetsierChatGPT2023} and likely has a large carbon footprint. Even without personalisation, \citet{stiennonLearning2020}'s RLHF model required 320 GPU days to train (p.8), suggesting the environmental impact of personalised LLM could be large.

\textit{\textbf{Environmental Harms}} are discussed by \citet{shelbyIdentifying2023} as a societal harm from damage to the natural environment and by \citet{weidingerTaxonomy2022} as environmental harms from operating LMs.

\section{A Three-Tiered Policy Framework for Personalised LLMs} \label{sec:policy_framework}
We propose a new policy framework for managing the benefits and risks of personalised LLMs. It provides a principled and holistic way of deciding how personalisation should be managed by different actors. 

\subsection{The Limits of Personalisation}
Deciding the limits of personalisation is inherently a normative decision, which involves making subjective and contentious choices about what should be permitted \cite{gabrielChallenge2021, kasirzadehConversation2022}. 
While it may be acceptable that a user wishes to interact with a \textit{rude} or \textit{sarcastic} personalised LLM, permitting users to create a \textit{racist} or \textit{extremist} model risks significant interpersonal and societal harms. 
Deciding the limits of personalisation straddles two separate issues: (i) deciding which \textit{aspects} of model behaviour should be personalised; and (ii) deciding \textit{how} should they be allowed to be personalised. 
For instance, it may be appropriate for personalised LLMs to express different views about political issues, but only within the limits of liberal democratic values. As such, facist beliefs would not be allowed, even though they are a political position.
Equally, it could be decided that some items should not be personalised at all. For instance, personalised LLMs might need to adhere to a minimum standard of safety when it comes to issues like threats of violence or child abuse. And, at the other end of the scale, complete freedom may be given over some attributes, such as the tone or style of outputs, which maximise utility and efficiency for the users of personalised LLM users and create few risks. 

A concrete way of addressing this issue is to think about the restrictions and requirements that should be applied to different types of personalised LLM outputs. By ``restrictions'' we mean things that the LLM application should not do; and by ``requirements'' we mean things that the LLM application should do. For instance, a restriction could be not allowing personalised models that produce hate speech when asked (e.g. ``Write something hateful against gay people''). A requirement could be that personalised models still have to adhere to a company's style guide; or to present multiple viewpoints on contentious topics (e.g. ``Should cannabis be legalised?''). 
In these cases, the restrictions and requirements present clear guardrails to ensure that, even when LLMs are personalised, they still operate within clearly specified boundaries. 
Note that both restrictions and requirements are normative ambitions; whether they can actually be implemented depends on the affordances of the technology in question (see \cref{sec:tech_challenges}).

\subsection{How People Interact with LLMs}
Workflows in machine learning models have changed dramatically over the past five years, which has affected how people will interact with personalised LLMs.
Most engineers do not train models from scratch. Instead they use pre-trained foundation models, which have been created by large AI labs, and then domain-adapt, fine-tune, prompt-tune or teach-in-context to create models for specific tasks \cite{bommasaniOpportunities2022}. For instance, generative models used for specific applications, such as to provide healthcare advice, might be fine-tuned on in-domain data assets, such as health information. 
Once models have been created, they are still only files and code. They need to be embedded within applications, which are what most users will actually interact with, like using ChatGPT via its interface.\footnote{\url{https://chat.openai.com/}. Note an API is now also available but most end-users will likely still interact with the interface due to lower barriers of entry.} 
Notwithstanding key differences across settings, the typical actors involved in creating an LLM application are as follows: 
\begin{itemize}
    \item The \textbf{model provider}. Refers to the organisation that makes the LLM available for use, typically through an API. The provider may or may not have built the entire model and/or may not be fully responsible for its development. Widely-known models include OpenAI's instruct-GPT3 \cite{ouyangTraining2022}, Anthropic's HHH assistants \cite{askellGeneral2021, baiConstitutional2022, baiConstitutional2022}, Google's LaMDA \cite{thoppilanLaMDA2022} and MetaAI's LLaMA \cite{metaaiIntroducing2023}.
    \item The \textbf{application provider}. Refers to the organisation that builds an application using the LLM. It can be provided through an API, interface or mobile app. For example, Jasper provide a copywriting service, AI21 provide a writing assistant,\footnote{\url{https://www.ai21.com/}} and OpenAI's provide a multi-purpose chatbot (ChatGPT). In some cases, the model provider and application provider will be the same actor.
    \item The \textbf{end-user}. Refers to the person who uses the interface, app or in some rare cases, may directly interact with an open-sourced model. This includes making requests, seeking advice, searching for information or otherwise chatting and interacting. 
    In principle, every internet user in the world can be an application user (depending on access constraints).
\end{itemize}

Machine learning workflows affect personalisation as, in principle, every actor involved in creating an LLM application could exert control over how it is adapted or personalised \cite{mokanderAuditing2023}. 
For example, if LLM creators do not impose any limits on personalisation then application providers would be free to adjust model behaviours as much as they like. This could lead to people using models which have very different political values from each other, which in turn might lead to social polarisation and division. 
Equally, application providers in a company may decide to give their staff full control over the outputs of a model; but this could create serious commercial risks if the staff chooses to write rude and abusive messages. 
However, at the same time, if LLM creators impose too many limits then application providers would not be able to meaningfully customise models, which would limit many of their benefits.
To ensure that the benefits of personalised LLMs are maintained, and the risks are mitigated, we need to ensure that freedom is maximised within the right limits: in practice, this means ensuring that decisions about personalisation are taken by appropriate actors.

\subsection{A Three-Tiered Policy Framework}
To help decide who should specify different policy items, we propose a three-tiered policy framework:
\begin{itemize}
    \item \textbf{\textit{Tier One}: Immutable restrictions}. Refers to types of model responses that must be restricted because they are very likely to be illegal at the national or supra-national level. 
    The specific restrictions will depend on the jurisdiction but will include terrorist content, written CSAM\footnote{Child Sexual Abuse Material.}, and language that threatens physical violence or sexual assault. 
    The \textbf{model provider} must implement the immutable restrictions.
    
    \item \textbf{\textit{Tier Two}: Optional restrictions and requirements}. Refers to types of model responses that are either required or restricted, based on the values and preferences of the actor releasing, controlling or hosting the LLM. Opted policies can be implemented by the \textbf{model provider} or the \textbf{application provider}.
    
    \item \textbf{\textit{Tier Three}: Tailored requirements}. Refers to types of model responses that the user wants to receive. 
    The \textbf{end-user} must decide their personal preferences for model responses, within the boundaries set at Tier One and Tier Two.
\end{itemize}

Policies in a higher tier cannot be violated by a policy in a lower tier. This means that a policy restricted at Tier One by the model provider cannot be overridden at Tier Two by the application provider, and a policy that is restricted at Tier Two cannot be overridden at Tier Three by the end-user. 
For instance, if an LLM is restricted from giving advice on bomb-making by the model provider at Tier One, then a user cannot be allowed to personalise the LLM at Tier Three to receive such advice. 
Model responses should be appropriate to the type of restriction or requirement that has been triggered at each tier. 
For instance, requests that trigger the immutable restrictions at Tier One should mostly be refused or blocked whilst requests that trigger the restrictions at Tier Two could be responded to with a more careful warning message, educational message, or by offering support.\footnote{Some technologies like DALLE-2 or ChatGPT already implement blocking of ``unsafe'' requests which violate the terms and conditions of OpenAI as the technology provider.}

The design of this framework gives more control to actors at the first tiers as the restrictions they impose cascade across all other actors.
For instance, if the creator of a widely-used LLM implements limits on how models can be personalised, it would affect every application provider which uses it. This is both an opportunity and a threat, if personalisation is not managed appropriately. 
Conceptually, this is analogous to Gillespie's idea of ``stacked moderation'' \cite{gillespieExpanding2020} whereby gatekeeping and infrastructural services for online platforms, such as hosting providers and app stores, can implicitly moderate those platforms by banning them or restricting their use. Although they do not directly affect the platforms' decisions, limiting reach and exposure is a powerful lever for change. Similarly, model providers have the potential to shape LLM personalisation and their decisions will constrain all other actors. 
\section{Discussion}
In this paper, we argue that the personalisation of LLMs is a likely pathway for the continued expansion in their deployment and public dissemination. To avoid a policy lag in understanding and governing LLMs, we attempt to document the landscape of personalised LLMs and their impacts now. We do so with two main contributions: (i) a taxonomy of the benefits and risks from personalised LLMs; and (ii) a policy framework to adequately govern these benefits and risks at three tiers of restrictions and requirements.

Throughout this work, we make the assumption that personalised LLMs are technically feasible with small advancements to current state of LLM technology; and that there will be a demand for and greater provision of personalisation in the near-future. We argue that these are realistic assumptions given the exhibited trend towards personalisation in other digital technologies; the existing implementation of the technical apparatus to adapt LLMs to human preferences via methods like RLHF; the apparent demand for increasingly customised and highly-adapted LLMs; and finally, the explicit plans from industry actors like OpenAI to grant users more flexibility in altering default model behaviours. However, the exact technical implementation of personalised LLMs is not pre-defined, and questions remain on how a model able to learn from personalised feedback could be implemented on the scale of a product like ChatGPT. We thus caveat our taxonomy by discussing some technical decision-points and engineering challenges which will impact the landscape of personalisation in LLMs (\cref{sec:tech_challenges}). We then discuss remaining challenges to implementing and enforcing a policy framework like ours (\cref{sec:policy_challenges}). Finally, we outline plans for maturing our research, and iterating on this first version of our taxonomy (\cref{sec:next_steps}).

\subsection{Technical Challenges}
\label{sec:tech_challenges}
We draw on some findings from the human feedback learning literature to hypothesise key design challenges and technical decision-points:

\paragraph{Data Efficiency and Quantity} Learning from human feedback is primarily applied during the fine-tuning stage so it relies on substantially less data to adapt model behaviour than pre-training. Most approaches train preference models on less than 50,000 datapoints \cite[e.g.][]{baiTraining2022, ouyangTraining2022}. A number of RLHF papers test performance over a range of data requirements \cite{perezDiscovering2022, stiennonLearning2020, zieglerFineTuning2019}. \citet{stiennonLearning2020}, for example, find that there are decreasing marginal returns to data scale in their reward model. The exact amount of data needed for effective personalisation in LLMs is unclear at present, but personalisation does introduce a new concern of how to adapt to new users without any feedback data points. This is commonly referred to as the cold start problem in recommender systems. Potential solutions already explored include batching of users into like-minded groups \cite{bakkerFinetuning2022} or recognising when a new user is similar to a known customisation case and then applying transfer learning \cite{moPersonalizing2016}. Insufficient personalised feedback data may hinder robustness -- with \citet{wangSelfInstruct2022} reporting a direct correlation between size and diversity of instructional data and the generalisability of models to unseen tasks; and \citet{bangEnabling2022} finding that their model does not generalise well to unseen human values. Potential solutions to reduce the amount of data needed include employing active learning techniques such as uncertainty sampling \cite{gaoAPRIL2018}; augmenting human-generated feedback data with synthetic data \cite{honovichUnnatural2022, wangSelfInstruct2022}; or adopting a rules-based approach so that users can define a set of guiding principles or ``constitutions'' \cite{baiConstitutional2022, glaeseImproving2022}. 

\paragraph{Data Format and Quality} Human preferences and values are inherently unstable and hard to precisely define \cite{gabrielChallenge2021}. Learning from revealed preferences over outputs is a way to optimise model performance under complexity of specifying a clear objective function \cite{zieglerFineTuning2019}. A wide variety of types of feedback data have been experimented with, including binary comparisons \cite{ganguliRed2022, zieglerFineTuning2019, jaquesHumancentric2020, baiConstitutional2022, askellGeneral2021}, ranked preferences \cite{askellGeneral2021, luBoosting2022}, demonstrations of optimal behaviours \cite{stiennonLearning2020, ouyangTraining2022, nakanoWebGPT2021} or revisions \cite{hancockLearning2019, liuSecond2023, wangNonParametric2021}. \citet{nguyenReinforcement2017} examine the robustness of reinforcement learning methods under more realistic properties of human feedback such as high variance, skew and restricted granularity, proposing an approach where performance does not degrade under noisy preference data. How much data needs to be collected depends on what data is collected: \citet{stiennonLearning2020} collect comparison data between the product of two documents, while \cite{nguyenMake2022} use a single document as input.

\paragraph{Model Efficiency and Training Complexity}
Beyond being data and labour intensive, adapting models to personalised human feedback may require substantial compute resources. Smaller yet more personalised models may be a preferred pathway because (i) model scale may not contribute significantly to performance \cite{stiennonLearning2020}, and (ii) increased scale may actually harm performance, for example leading to increased sycophancy or goal preservation \cite{perezDiscovering2022}. A number of works point to the competitiveness of rejection sampling at inference time instead of applying the full RLHF fine-tuning pipeline \cite[e.g. see][]{xuLearning2022, bakkerFinetuning2022}. Training complexity could be further reduced by implementing batched or offline training \cite{stiennonLearning2020, nguyenMake2022, liuAligning2022}, instead of online training \cite{zieglerFineTuning2019}.

\paragraph{Alignment Tax} A concern with RLHF techniques is model overfitting \cite{baiTraining2022}. Thus, any approach to personalisation must carefully balance effects on performance from degraded language representation -- the so called ``alignment tax''. However, many works have demonstrated little to no alignment tax \cite{glaeseImproving2022, askellGeneral2021, baiTraining2022, liuAligning2022}. Often a KL-divergence penalty is included during training to prevent the fine-tuned model deviating too far from the pre-trained representations \cite{zieglerFineTuning2019, liuAligning2022, ouyangTraining2022, nakanoWebGPT2021, baiTraining2022}.

\paragraph{Interpretability and Oversight} Larger and more powerful LLMs, that are adapted to increasingly complex tasks, pose a challenge for effective oversight from their end-users or technology providers \cite{wuRecursively2021, leikeScalable2018, bowmanMeasuring2022}. As \citet{ouyangTraining2022} note ``one of the biggest open questions is how to design an alignment process that is transparent'' (p.19). A concern with personalised models is that their behaviour may be less interpretable and transparent, for example due to multiple branches or versions of the model. This has implications for AI safety and effective evaluation of model behaviours. Combining preference reward modelling with a rules-based reward model is a promising solution for controlling model behaviour \cite{glaeseImproving2022}, as well as defining behaviours via a ``constitution'', which \citet{baiConstitutional2022} argue is ``a simple and transparent form'' to encode and evaluate desirable behaviours.

\subsection{Policy Framework Enforcement}
\label{sec:policy_challenges}
Our taxonomy demonstrates that personalised LLMs could have wide-reaching benefits, but also come with a set of concerning risks. In order to balance these benefits with risks, we provide a framework which outlines some properties for the appropriate governance of personalised LLMs. The goal of policy enforcement is to ensure that violations are minimised, particularly where there is a serious risk of harm; and that the friction applied to users' experiences is proportionate. However, enforcement of the policy framework will always be imperfect, as will enforcement of specific requirements and restrictions at each tier. Some specific challenges remain:

\paragraph{Compliance with Existing Regulation} Any technology permitting the personalisation of LLMs would need to comply with existing regulations and laws such as the GDPR \cite{europeanparliamentGDPR2016}, the Online Safety Bill \cite{ukparliamentOnline2023} and the various European AI standards \cite{commissionArtificial2021, commissionAI2022}. It is harder to assess whether an LLM is ``fair'' or ``harmful'' to its user when the space for possible personalisation actions is so complex. These issues also apply to LLMs in general, where adaptation of pre-trained models to downstream applications pose significant challenges to traditional auditing approaches \cite{mokanderAuditing2023}.

\paragraph{Defining New Regulation and Oversight} It is challenging to regulate or evaluate a system which is both \textit{dynamic} (adapting continually or frequently to user feedback) and \textit{distributed} (personalised across many users). We aim to address this issue by imposing some constraints which are defined and implemented at a high level (Tier One) and stable through time, i.e., applying across all users and all training updates. However, as model behaviours shift in response to specific user feedback, it could become harder to monitor model behaviours and maintain effective oversight.

\paragraph{Distributed Responsibility} Our framework distributes responsibility among model providers, application providers and end-users, with the first two of these actors bearing the brunt of defining the \textit{bounds} of personalisation. We at best assume that these are ``good faith'' actors who are incentivised (via profit or user footfall) to responsibly balance personalisation with adequate safeguards from harm; and at worst assume that they will be regulated to do so. In reality, with models like LLaMA \cite{metaaiIntroducing2023} being open-sourced, anyone with sufficient technical skill and compute resources could hypothetically create, launch and host a system capable of personalisation. Such fragmentation of the development landscape in LLMs poses a challenge for audit and policy enforcement \cite{mokanderAuditing2023}.

\paragraph{Populating Tiers of the Policy} While we define a clear functional framework, we have not fully specified principles that determine which restrictions or requirements fall under each tier. For example, why some risks come under Tier One (and are never allowed) while others are optionally defined in Tier Two. This is a problem also faced by online trust and safety regulation -- where for example, early iterations of the Online Safety Bill \cite{ukparliamentOnline2023} included separate treatment of \textit{illegal} content versus \textit{legal but harmful} content, as well as tiered restrictions for children and minors versus adults. It is clear that any content which already violates existing laws and regulations in the operating jurisdiction, such as hate speech, CSAM or terrorist content, would inherently need to be restricted in Tier One. Beyond this, we leave the open question of what principles could be employed by model or application providers to define their own organisational bounds.

\subsection{Next Steps}
\label{sec:next_steps}
A large amount of our work in this paper, both in the taxonomy and in the policy framework, relies on informed speculation over the future development and governance of personalised LLMs. The affordances, constraints and harms from any technology depend critically on how it is designed, how its outputs are used in the real world and what safeguards or regulations are provisioned to guide its impact post-deployment. None of these conditions are presently clear; so, we very much consider this to be a first version of our work. In the future, we plan to iterate on the taxonomy and policy framework by conducting semi-structured interviews with (i) the end-users of LLMs to understand their priorities and concerns; (ii) model and application providers to scope their desires for enabling personalisation and for defining limits, as well as the technical apparatus available to them to both of these things; and finally, (iii) policymakers to assess how personalised LLMs fit into existing laws and regulations, and the new challenges they pose. By starting the conversation now, we hope to avoid long lags in understanding, documenting and governing the harms from personalised LLMs as a future technology which could widely impact individuals and society. 

\section*{Acknowledgements} This paper, which began as an exploration of how to improve feedback between humans-and-model-in-the-loop, is part of a body of work funded by a MetaAI Dynabench grant. H.R.K's PhD is supported by the Economic and Social Research Council grant ES/P000649/1. P.R's PhD is supported by the German Academic Scholarship Foundation. We particularly want to thank Andrew Bean for his thoughtful input and assistance with the literature review. This paper is also the product of many interesting conversations at various conferences and working groups. We hope to survey the opinions and feedback of many other stakeholders in the future (including end-users, policy makers and technology providers) to further enrich our discussion. 

\bibliographystyle{acl_natbib}
\bibliography{language-agent-alignment}


\end{document}